\documentclass[a4paper]{spie} 

\usepackage{amsmath,amsfonts,amssymb}
\usepackage{graphicx}
\usepackage{hyperref}
\usepackage[margin=2cm]{geometry}
\usepackage{subcaption}
\usepackage{array, booktabs, siunitx}
\hypersetup{colorlinks=true,linkcolor=[rgb]{0.1,0.3,0.9},urlcolor=[rgb]{0.2,0.2,0.2},citecolor=[rgb]{0.1,0.3,0.9}}

\usepackage{array, booktabs, siunitx}
\usepackage[table]{xcolor}

\title{MRI Embeddings Complement Clinical Predictors for Cognitive Decline Modeling in Alzheimer’s Disease Cohorts}

\author[]{Nathaniel Putera}
\author[]{Daniel Vilet Rodríguez}
\author[]{Noah Videcrantz}
\author[]{\\\vspace{-0.9em} Julia Machnio}
\author[]{Mostafa Mehdipour Ghazi}
\affil[]{Department of Computer Science, University of Copenhagen, Copenhagen, Denmark}

\begin{document}
\maketitle

\begin{abstract}
Accurate modeling of cognitive decline in Alzheimer's disease is essential for early stratification and personalized management. While tabular predictors provide robust markers of global risk, their ability to capture subtle brain changes remains limited. In this study, we evaluate the predictive contributions of tabular and imaging-based representations, with a focus on transformer-derived Magnetic Resonance Imaging (MRI) embeddings. We introduce a trajectory-aware labeling strategy based on Dynamic Time Warping clustering to capture heterogeneous patterns of cognitive change, and train a 3D Vision Transformer (ViT) via unsupervised reconstruction on harmonized and augmented MRI data to obtain anatomy-preserving embeddings without progression labels. The pretrained encoder embeddings are subsequently assessed using both traditional machine learning classifiers and deep learning heads, and compared against tabular representations and convolutional network baselines. Results highlight complementary strengths across modalities. Clinical and volumetric features achieved the highest AUCs of around 0.70 for predicting mild and severe progression, underscoring their utility in capturing global decline trajectories. In contrast, MRI embeddings from the ViT model were most effective in distinguishing cognitively stable individuals with an AUC of 0.71. However, all approaches struggled in the heterogeneous moderate group. These findings indicate that clinical features excel in identifying high-risk extremes, whereas transformer-based MRI embeddings are more sensitive to subtle markers of stability, motivating multimodal fusion strategies for AD progression modeling.
\end{abstract}


\section{Introduction}

Cognitive decline is an indicator of Alzheimer’s disease (AD), a progressive neurodegenerative disorder affecting memory, language, and executive function [\citenum{better2023alzheimer}]. As the most common cause of dementia worldwide [\citenum{gustavsson_global_2023}], AD imposes not only individual suffering but also substantial societal and economic burdens through increased care needs and loss of independence. Yet, its clinical course is highly variable; some individuals decline gradually over many years, while others progress rapidly to severe impairment. This heterogeneity complicates prognosis and care planning. Identifying individuals at risk of faster decline is thus essential for enabling timely intervention, optimizing clinical trial design, and advancing personalized treatment strategies [\citenum{yi2023identifying}].

Conventional diagnosis and staging of AD rely on neuropsychological tests, structural magnetic resonance imaging (MRI), and fluid or molecular biomarkers [\citenum{jack_jr_revised_2024}]. Structural MRI is widely available and non-invasive, and it captures atrophy patterns associated with disease progression [\citenum{sperling_toward_2011}]. However, cross-sectional measures such as hippocampal volume provide only limited information on the temporal dynamics of decline. This limitation has motivated the development of longitudinal methods that characterize trajectories of cognitive changes.

Recent advances in deep learning have enabled data-driven modeling of disease trajectories from neuroimaging [\citenum{yi2023identifying,helaly2022deep}]. Nevertheless, most studies remain focused on static classification of AD stages rather than predicting future decline [\citenum{chamakuri2025systematic}]. Directly training large models on MRI data also risks overfitting in modest cohort sizes [\citenum{jo2019deep}]. To mitigate this, unsupervised and self-supervised learning strategies are increasingly employed to extract robust, transferable representations of brain structure without requiring large labeled datasets [\citenum{chen_vit-v-net_2021}].

In this work, we aim to predict cognitive decline trajectories in AD by utilizing baseline MRIs and tabular descriptors. We derive four progression groups of stable, mild, moderate, and severe decline, based on clustering of temporal Clinical Dementia Rating Sum of Boxes (CDR-SB) scores. Besides, we train a 3D vision transformer (ViT) inspired by [\citenum{chen_vit-v-net_2021}] using MRI reconstruction as a pretext task, enabling the extraction of meaningful embeddings independent of clinical labels. These representations are subsequently used for downstream classification and compared against traditional classifiers, tabular predictors, and pretrained convolutional networks.

Our work provides three main contributions. First, we derive progression labels using trajectory clustering with Dynamic Time Warping (DTW) to capture heterogeneous temporal patterns despite irregular follow-up schedules. Second, we train a 3D ViT model through unsupervised MRI reconstruction to obtain anatomy-aware embeddings, thereby eliminating reliance on clinical labels. Third, we perform a systematic cross-modal comparison between tabular predictors, transformer-derived imaging features, and convolutional network baselines to characterize modality-specific strengths in predicting distinct progression subtypes.

\section{Methods}

\subsection{Study Data}

We used two complementary subsets from the Alzheimer’s Disease Neuroimaging Initiative (ADNI) [\citenum{wyman2013standardization}]. The first subset from ADNI1 consists of screening-visit data with 503 T1 MRI scans. These scans were employed to train the vision encoder and support image-based predictive modeling. All MRIs underwent preprocessing with the FreeSurfer pipeline [\citenum{fischl2002whole}], including motion and bias-field correction, affine registration to the MNI305 template space, resampling to $1$ mm isotropic resolution, and intensity normalization to $[0, 255]$ [\citenum{nielsen2025assessing}]. To ensure fair representation across patient groups, the MRI subset was first stratified by age, sex, and clinical diagnosis. The stratified data were then divided into training and testing sets using an 80:20 split. We applied a series of data augmentation techniques to increase variability and mitigate overfitting, following the protocol of [\citenum{mehdipour2025fast}]. These transformations simulate common clinical MRI artifacts as well as natural anatomical variability, thereby encouraging the model to learn robust features that generalize beyond the training data.

The second subset, ADNI-Merge [\citenum{mehdipour2024comparative}], provides longitudinal tabular information, including cerebrospinal fluid (CSF) biomarkers (amyloid-beta, tau, p-tau), positron emission tomography (PET) measurements (FDG, AV45, PIB, FBB), risk factors (age, sex, education, APOE4), neuropsychological test scores (CDR, ADAS, MMSE, RAVLT, LDEL, DIGIT, TRAB, FAQ, MOCA), and brain volumetrics (hippocampus, entorhinal, middle temporal, fusiform, ventricles). From this subset, we derived cognitive progression labels using the CDR-SB. Specifically, we applied trajectory clustering to assign individuals to distinct progression groups that capture the heterogeneity of cognitive decline. The resulting labels served as ground truth for downstream prediction, while the remaining imaging and tabular data were used as predictors in classification experiments.

\subsection{Cognitive Decline Groups}

\begin{figure}[t]
\centering
\begin{subfigure}{0.5\textwidth}
\centering
\includegraphics[width=\linewidth]{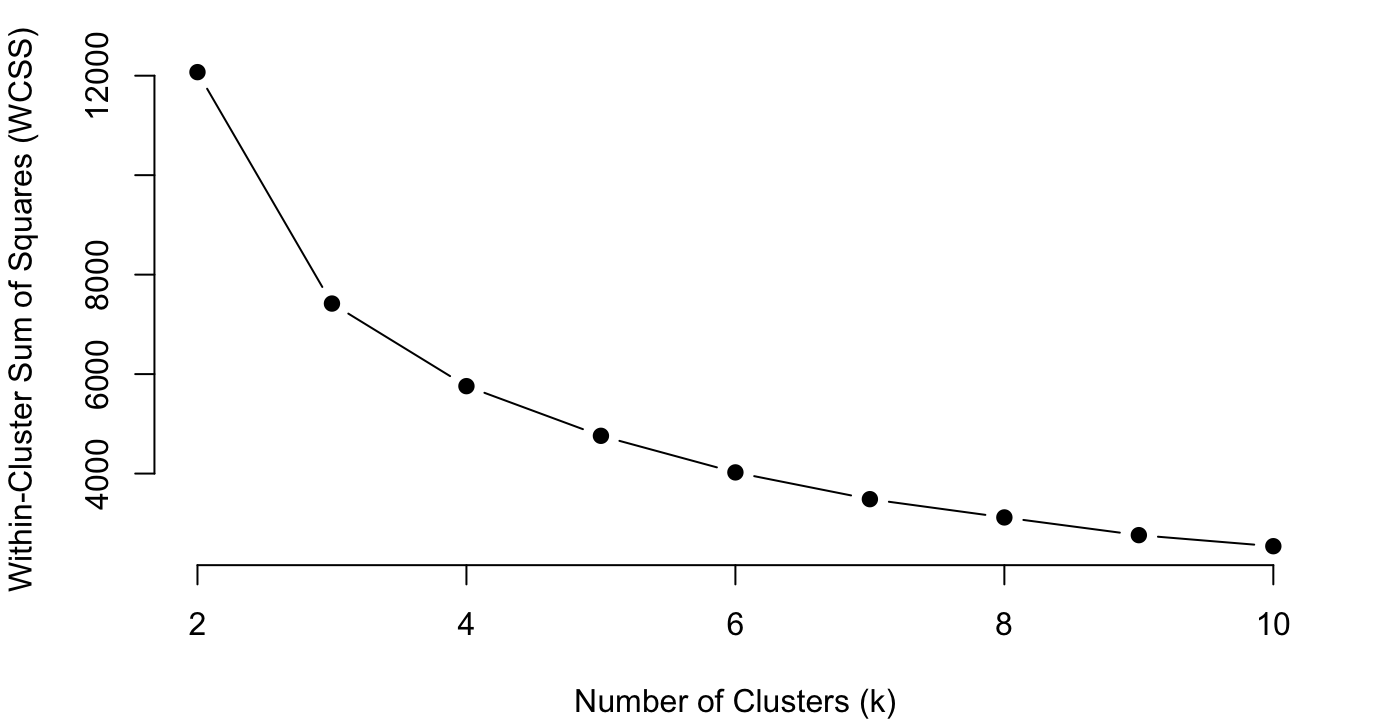}
\end{subfigure}%
\hspace{1em}
\begin{subfigure}{0.35\textwidth}
\centering
\includegraphics[width=\linewidth]{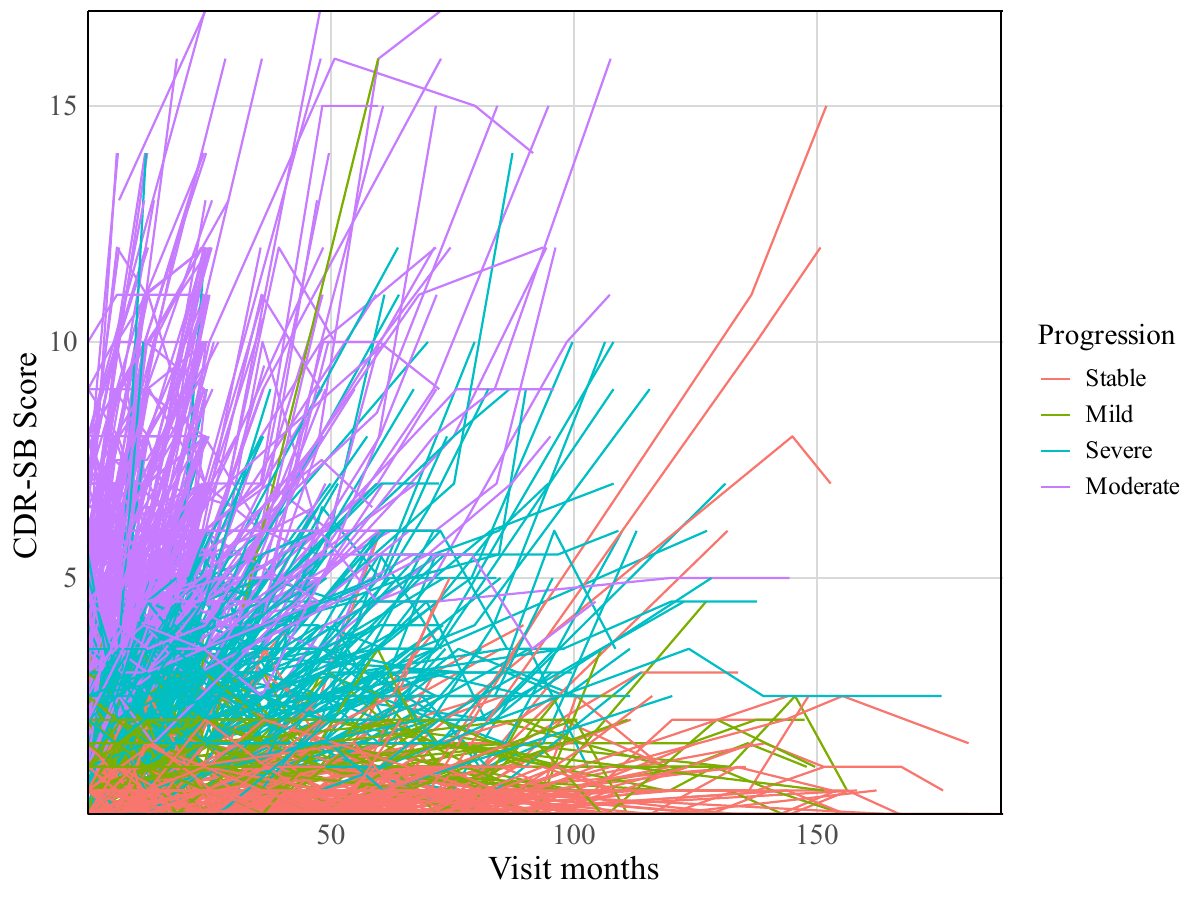}
\end{subfigure}
\caption{Elbow plot for selecting the optimal number of clusters (left), and the resulting 4 clusters (right) obtained from $k$-means clustering with DTW distances, grouped by CDR-SB progression.}
\label{AgeGroup_stats}
\end{figure}

Cognitive decline groups were defined using the CDR-SB scores from ADNI-Merge participants with at least two visits. It was chosen because it is a well-established measure of functional and cognitive status in AD, widely used in clinical trials and observational studies [\citenum{cedarbaum2013rationale}]. To capture temporal similarities across individual trajectories, we applied $k$-means clustering using DTW distances [\citenum{ghazi2024cognitive}]. DTW enables alignment of trajectories of varying length and timing, allowing us to identify homogeneous patterns of change even when visits occur at different intervals. We set $k=4$ to define four clinically interpretable subgroups. These were labeled as stable, mild, moderate, and severe, corresponding to increasing levels of cognitive decline over time. These clusters served as target labels for downstream predictive modeling. An overview of the clustered trajectories is shown in Figure~\ref{AgeGroup_stats}, indicating the distinct CDR-SB progression profiles across groups.

\subsection{3D ViT Pretraining}

To obtain robust MRI-based embeddings while reducing the risk of overfitting associated with direct supervised training, we adopted an unsupervised pretraining strategy. Specifically, we implemented a 3D ViT-V-Net model inspired by [\citenum{chen_vit-v-net_2021}], designed to learn representations through image reconstruction. The model was trained for 200 epochs with a mean squared error (MSE) loss, using both original and artifact-simulated MRI scans. Reconstruction quality was monitored with the Structural Similarity Index Measure (SSIM) to ensure that the learned representations preserved fine-grained anatomical structure.

Following pretraining, the decoder was discarded, and the encoder was retained to extract latent feature embeddings from MRI scans. These embeddings served as input representations for downstream classification tasks, where we evaluated both traditional machine learning classifiers and fully connected (FC) classifier heads. To provide a comparative baseline, we fine-tuned ResNet-3D models [\citenum{chen2019med3d,cardoso_monai_2022}] on the same MRI subset, enabling a direct assessment of performance differences between transformer-based and convolutional architectures.

\subsection{Cognitive Decline Prediction}

To evaluate predictive strategies for AD progression, we considered models based on tabular predictors, transformer embeddings, and pretrained convolutional neural networks (CNNs). All approaches were trained to classify individuals into the four cognitive decline groups defined by clustered CDR-SB trajectories.

\paragraph{\textbf{Tabular Predictors.}}

Tabular features included CSF markers, PET measures, brain volumetrics, neuropsychological test scores (excluding CDR-SB), and risk factors. To obtain compact and noise-robust representations, each feature set was processed through an autoencoder before classification. The resulting representations were then used to train an Extreme Gradient Boosting (XGBoost) model [\citenum{ali2023extreme}] for the four-class prediction task.

\paragraph{\textbf{3D ViT Predictors.}}

MRI-based representations were extracted from the encoder of the pretrained ViT-V-Net model. To reduce dimensionality and mitigate overfitting, we applied principal component analysis (PCA), retaining 15 principal components that together explained approximately 95\% of the variance. These reduced features were classified with XGBoost. In a complementary setup, we attached a lightweight FC head to the frozen ViT-V-Net encoder, consisting of layers with ReLU activations and dropout for regularization. Only the classifier head was trained, using a learning rate of $1\times10^{-4}$.

\paragraph{\textbf{3D CNN Predictors.}}

For comparison with established CNN architectures, we fine-tuned pretrained 3D models. In the ResNet-Med3D framework [\citenum{chen2019med3d}], the original segmentation head was replaced by a custom classifier. The backbone was updated with a learning rate of $1\times10^{-4}$, while the new classifier head was trained with $1\times10^{-2}$. Intensity normalization was applied to match the input distribution of the pretraining model. Similarly, in the ResNet-MONAI implementation [\citenum{cardoso_monai_2022}], the final output layer of the backbone model was adapted to predict the four cognitive decline categories.

\section{Results}

We evaluated four-way classification of cognitive decline trajectories using one-vs-rest AUC as the primary metric. Summary AUCs for tabular predictors are reported in Table~\ref{tab:auc_results}, and a per-class breakdown across all other methods is provided in Table~\ref{tab:auc-breakdown}. As can be seen, using clinical and volumetric features compressed via autoencoders, with XGBoost yielded the strongest performance for the mild and severe groups, indicating that structured clinical information and brain volumetrics are particularly informative for progressive groups.

We next assessed MRI-derived embeddings from the pretrained ViT-V-Net encoder combined with PCA and XGBoost. Relative to tabular models, this approach improved discrimination of the stable group, suggesting that the learned anatomical representations capture features associated with minimal near-term change. However, performance in the severe and mild groups did not surpass the tabular baseline, potentially reflecting information loss from dimensionality reduction or the need for features more tightly coupled to disease severity.

Attaching a lightweight FC head to the frozen ViT-V-Net encoder produced the highest AUC for the stable class among all imaging models. Despite these gains, the moderate class remained the most challenging across methods, reinforcing that borderline progression states are difficult to separate from baseline MRI alone. Moreover, among CNN baselines, ResNet-Med3D outperformed the ResNet-MONAI, but both CNN variants underperformed the ViT-based models on the stable class and showed limited improvement for moderate and severe trajectories. Overall, tabular predictors excel at identifying mild and severe decline, whereas transformer-based MRI representations are most effective for detecting stable courses. This complementary pattern motivates multimodal fusion and class-aware learning in future work to leverage clinical severity cues alongside anatomy-driven stability signatures for more robust AD progression modeling.

\begin{table}[t]
\centering
\caption{Test AUCs (mean $\pm$ std) of different feature sets for predicting AD progression stages over 5 repetitions.}
\vspace{0.1cm}
\renewcommand{\arraystretch}{1.0}
\setlength{\tabcolsep}{8pt}
\begin{tabular}{lcccc}
\toprule
\textbf{Feature Set} & \textbf{Stable} & \textbf{Mild} & \textbf{Moderate} & \textbf{Severe} \\ 
\midrule
Cognitive Scores  & \textbf{0.59} ± 0.03  & \textbf{0.75} ± 0.01 & \underline{0.56} ± 0.03 & \textbf{0.77} ± 0.02  \\
\midrule
CSF Markers & 0.55 ± 0.02 & 0.64 ± 0.02 & \textbf{0.59} ± 0.02 & 0.59 ± 0.01 \\
PET Measures & 0.49 ± 0.02 & 0.69 ± 0.02  & \textbf{0.59} ± 0.02 & 0.69 ± 0.02 \\
Risk Factors & 0.53 ± 0.01 & 0.63 ± 0.02 & 0.51 ± 0.02 & 0.62 ± 0.02 \\
Brain Volumetrics & \underline{0.58} ± 0.03 & \underline{0.70} ± 0.02 & 0.53 ± 0.03 & \underline{0.70} ± 0.02 \\
\bottomrule
\end{tabular}
\label{tab:auc_results}
\vspace{-0.1cm}
\end{table}

\begin{table}[t]
\vspace{0.5cm}
\centering
\caption{Test AUCs (mean $\pm$ std) of different methods for predicting AD progression stages over 5 repetitions.}
\vspace{0.1cm}
\renewcommand{\arraystretch}{1.0}
\setlength{\tabcolsep}{8pt}
\begin{tabular}{p{4cm}cccc}
\toprule
\textbf{Method} & \textbf{Stable} & \textbf{Mild} & \textbf{Moderate} & \textbf{Severe} \\ 
\midrule
Brain Volumetrics & 0.58 ± 0.03 & \textbf{0.70} ± 0.02 & \textbf{0.53} ± 0.03 & \textbf{0.70} ± 0.02 \\  
ViT + XGBoost & \underline{0.65} ± 0.02 & \underline{0.59} ± 0.02 & \underline{0.51} ± 0.03 & 0.46 ± 0.02 \\  
ViT + FC & \textbf{0.71} ± 0.01 & 0.58 ± 0.03 & 0.37 ± 0.02 & 0.41 ± 0.02 \\
ResNet-Med3D & 0.62 ± 0.02 & 0.53 ± 0.02 & 0.43 ± 0.02  & \underline{0.54} ± 0.01 \\ 
ResNet-MONAI & 0.55 ± 0.03 & 0.52 ± 0.02 & 0.47 ± 0.02 & 0.48 ± 0.02 \\ 
\bottomrule
\end{tabular}
\label{tab:auc-breakdown}
\vspace{-0.1cm}
\end{table}

\section{Conclusion}

This study investigated how different data modalities and representation learning strategies contribute to the prediction of cognitive decline in AD progression. We first applied stratified preprocessing and realistic data augmentation to ensure balanced group representation and robustness to common MRI artifacts. To obtain meaningful imaging features, we pretrained a 3D ViT model for unsupervised reconstruction, retaining the encoder as a feature extractor for downstream tasks. The resulting embeddings were compared against clinical predictors, volumetric measures, and CNN architectures.

Our results demonstrated complementary strengths of different modalities. Tabular predictors were most effective in identifying mild and severe progression groups, achieving average AUC values of approximately 0.70. In contrast, MRI-derived embeddings from ViT were most informative for the stable progression group, achieving the highest AUC of 0.71. The difficulty in classifying the moderate group highlights the clinical and biological heterogeneity of this stage, which may require multimodal integration for more reliable discrimination.

Overall, clinical features capture global risk patterns, while deep learning–based imaging embeddings characterize subtle structural signatures of stability and transition. The complementary performance profiles indicate that future work should move toward multimodal fusion strategies, combining structured tabular data with transformer-derived imaging features. Such integrative approaches hold promise for more robust AD progression modeling, ultimately supporting early stratification and personalized disease management.

\section*{ACKNOWLEDGMENTS}       

This project is supported by the Pioneer Centre for AI, funded by the Danish National Research Foundation (grant number P1).

\bibliographystyle{spiebib}
\bibliography{references}

\end{document}